# Face Recognition Based on SVM and 2DPCA

Thai Hoang Le, Len Bui

*Faculty of Information Technology, HCMC University of Science*
*Faculty of Information Sciences and Engineering, University of Canberra*
lhthai@fit.hcmus.edu.vn, len.bui@canberra.edu.au

*Abstract*

*The paper will present a novel approach for solving face recognition problem. Our method combines 2D Principal Component Analysis (2DPCA), one of the prominent methods for extracting feature vectors, and Support Vector Machine (SVM), the most powerful discriminative method for classification. Experiments based on proposed method have been conducted on two public data sets FERET and AT&T; the results show that the proposed method could improve the classification rates.*

## 1. Introduction

Human faces contain a lot of important biometric information. The information can be used in a variety of civilian and law enforcement applications. For example, identity verification for physical access control in buildings or security areas is one of the most common face recognition applications. At the access point, an image of a claimed person's face is captured by a camera and is compared with stored images of the claimed persons. Then it will be accepted only if it is matched. For high security areas, a combination with card terminals is possible, so that a double check is performed.

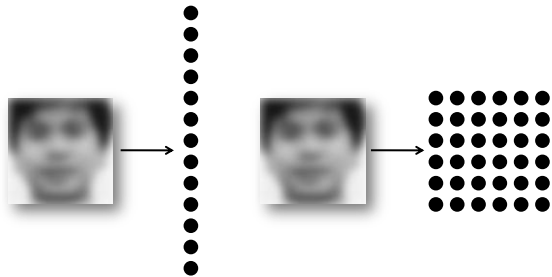

**Figure 1. Image Representations in PCA and 2DPCA**

Since Matthew Turk and Alex Pentland [1] used Principal Component Analysis (PCA) to deal with the face recognition problem, PCA has become the standard method to extract feature vectors in face recognition because it is stable and has good performance. Nevertheless, PCA could not capture all local variances of images unless this information is explicitly provided in the training data. To deal with this problem, some researchers proposed other approaches. For example, Wiskott et al. [2] suggested a technique known as elastic bunch graph matching to extract local features of face images. Penev and Atick [3] proposed using local features to represent faces; they used PCA to extract local feature vectors. They reported that there was a significant improvement in face recognition. Bartlett et al. [4] proposed using independent component analysis (ICA) for face representation to extract higher dependents of face images that cannot represented by Gaussian distributions, and reported that it performed better than PCA. Ming-HsuanYang [5] suggested Kernel PCA (or nonlinear subspace) for face feature extraction and recognition and described that his method outperformed PCA (linear subspace). However, the performance costs of them are higher than PCA.

To solve these problems, Jian Yang [6] proposed a new method called 2D Principal Component Analysis (2DPCA). In conventional PCA, face images have been represented in vectors by some technique like concatenation. As opposed to PCA, 2DPCA represents face images by using matrices or 2D images instead of vectors (Fig. 1). Clearly, using 2D images directly is quite simple and local information of the original images is preserved sufficiently, which may bring more important features for facial representation. In face identification, some face images are easy to recognize, but others are hard to identify; for example, frontal face images are easier than to be recognized than profile face images. Therefore, we proposed a weighted-2DPCA model to deal with the difficulty.

In 1995, Vapnik and Cortes [7] presented the foundations for Support Vector Machine (SVM). Since then, it has become the prominent method to solve problems in pattern classification and regression. The basic idea behind SVM is finding the optimal linear hyperplane such that the expected classification error for future test samples is minimized, i.e., good generalization performance. Obviously, the goal of all classifiers is not to get the lowest training error. For example, a k-NN classifier can achieve the accuracy rate 100% with k=1. However, in practice, it is the worst classifier because it has high structural risk.

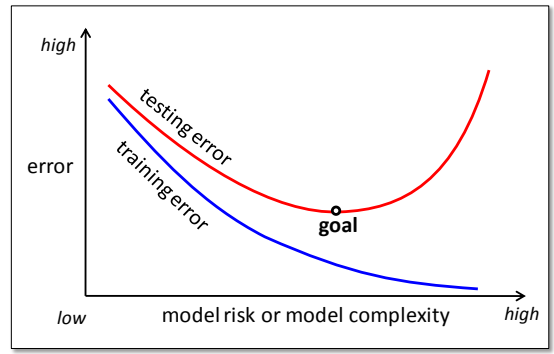

**Figure 2. Curves of Testing Error and Training Error**

They suggested the formula testing error = training error + risk of model (Fig. 2). To achieve the goal to get the lowest testing error, they proposed the structural risk minimization inductive principle. It means that a discriminative function that classifies the training data accurately and belongs to a set of functions with the lowest VC dimension will generalize best results regardless of the dimensionality of the input space. Based on this principle, an optimal linear discriminative function has been found. For linearly non-separable data, SVM maps the input to a higher dimensional feature space where a linear hyperplane can be found. Although there is no warranty that a linear solution will always exist in the higher dimensional space, it is able to find effective solutions in practice. To deal with the face gender classification, many researchers [8-11] have applied SVM in their studies and stated that the experiment results are very positive. In our research, we have combined the power of each method, weighted-2DPCA and SVM, to solve the problem.

The remaining sections of our paper will discuss the implementation of our face recognition system, related theory, and experiments. Section 2 gives details of 2DPCA. Section 3 discusses how to use SVM in face classification. In Section 4, we will describe the implementation and experiments. Finally, Section 5 is our conclusion.

## 2. 2D Principal Component Analysis

### 2.1. Face Model Construction

As mentioned above, we propose a weighed-2DPCA to deal with some practical situations in which some face images in database are difficult to identify due to their poses (front or profile) or their qualities (noise, blur).

Training data $D = \{(\mathbf{A}^{(i)}, w_i), i = 1,...,N\}$

Algorithm 1: Construct proposed face model
Step 1: Compute the mean image

$$\overline{\mathbf{A}} = \frac{\sum_{i=1}^{N} w_i \mathbf{A}^{(i)}}{\sum_{i=1}^{N} w_i} \quad (1)$$

Step 2: Compute matrix

$$\mathbf{G} = \frac{\sum_{i=1}^{N} w_i (\mathbf{A}^{(i)} - \overline{\mathbf{A}})^T (\mathbf{A}^{(i)} - \overline{\mathbf{A}})}{\sum_{i=1}^{N} w_i} \quad (2)$$

Step 3: Compute eigenvectors $\{\mathbf{\Omega}_1, \mathbf{\Omega}_2,...,\mathbf{\Omega}_n\}$ and eigenvalues $\{\lambda_1, \lambda_2,...,\lambda_n\}$ of G.

### 2.2. Feature Extraction

First, a projection point of image A on 2DPCA space is matrix $(\mathbf{X}_1, \mathbf{X}_2,...,\mathbf{X}_n)$

$$\mathbf{X}_k = (\mathbf{A} - \overline{\mathbf{A}})\mathbf{\Omega}_k, k = 1,...,d \quad (3)$$

Second, the matrix is projected on PCA space to convert matrix to vector and reduce the dimension.

## 3. Support Vector Machine

The goal of SVM classifiers is to find a hyperplane that separates the largest fraction of a labeled data set $\{(\mathbf{x}^{(i)}, y^{(i)}); \mathbf{x}^{(i)} \in \Box^n; y^{(i)} \in \{-1,+1\}; i = 1,...,N\}$. The most important requirement, which the classifiers must have, is that it has to maximize the distance or the margin between each class and the hyperplane (Fig 3.).

In most of real applications, the data could not be linearly classified. To deal with this problem, we transform data into a higher dimensional feature space and assume that our data in this space can be linearly classified (See Fig 4.).

$$\Phi : \square^n \to \square^m$$
$$\mathbf{x} \mapsto \Phi(\mathbf{x})$$
(4)

In fact, determining the optimal hyperplane is a constrained optimization problem and can be solved using quadratic programming techniques. The discriminant hyperplane is defined as the following

$$y(\mathbf{x}) = \sum_{i=1}^{N} \alpha_i y^{(i)} K(\mathbf{x}^{(i)}, \mathbf{x}) + b \tag{5}$$

where $K(\mathbf{x}', \mathbf{x}'')$ is the kernel function.

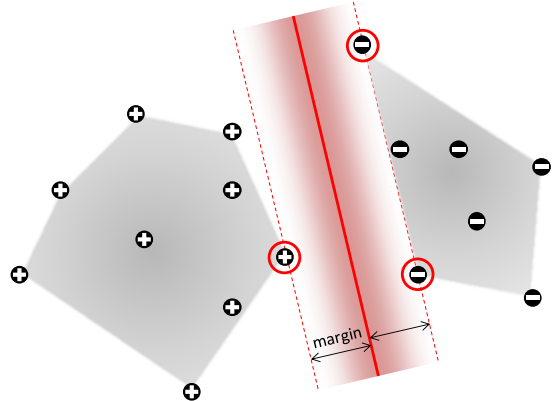

**Figure 3. An SVM Classifier**

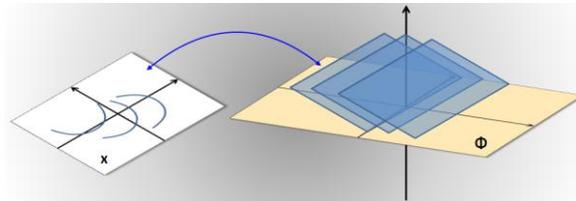

**Figure 4. Input Space and Feature Space**

### 3.1. Classifier Construction Phase

Algorithm 2: Construct classifier

Step 1: Compute matrix H

$$H_{ij} = y^{(i)} y^{(j)} K(\mathbf{x}^{(i)}, \mathbf{x}^{(j)}) \tag{6}$$

Step 2: Use quadratic solver to solve the optimization problem with objective function:

$$\boldsymbol{\alpha} = \underset{\boldsymbol{\alpha}}{\operatorname{argmin}} \left( \frac{1}{2} \boldsymbol{\alpha}^T \mathbf{H} \boldsymbol{\alpha} - \sum_{i=1}^{N} \alpha_i \right)$$
$$\begin{cases} 0 \le \alpha_i \le C \\ \sum_{i=1}^{N} \alpha_i y^{(i)} = 0 \end{cases} \tag{7}$$

Step 3: Compute b

$$idx = \{i \mid \alpha_i > 0\}$$
$$N_{idx} = |idx|$$
$$b = \frac{1}{N_{idx}} \sum_{i \in idx} \left( y^{(i)} - \sum_{j \in idx} \alpha_j y^{(j)} K\left(\mathbf{x}^{(j)}, \mathbf{x}^{(i)}\right) \right) \quad (8)$$

### 3.2. Classification Phase

Algorithm 3: Classify

Step 1: Compute the value y

$$y = \text{sgn}\left( \sum_{i=1}^{N} \alpha_i y^{(i)} K\left(\mathbf{x}^{(i)}, \mathbf{x}\right) + b \right) \quad (9)$$

Step 2: Classify for x

$$\begin{cases} \text{if } y = +1 \text{ then } \mathbf{x} \text{ belong class } \{+1\} \\ \text{if } y = -1 \text{ then } \mathbf{x} \text{ belong class } \{-1\} \end{cases} \quad (10)$$

### 3.3. SVM for Face Identification

To apply SVM in face recognition, we use One-Against-All decomposition to transform multi-class problem to a set of two-class problems.

Training set $D = \{(\mathbf{x}^{(i)}, y^{(i)}); \mathbf{x}^{(i)} \in \square^n; y^{(i)} \in \{-1, +1\}; i = 1, ..., N\}$ is transformed to series of $D_k = \{(\mathbf{x}^{(i)}, y_k^{(i)}); y_k^{(i)} \in \{-1, +1\}\}$

where

$$y_k^{(i)} = \begin{cases} +1 & y^{(i)} = k \\ -1 & y^{(i)} \neq k \end{cases} \quad (11)$$

Algorithm 2 is used to compute the discriminant functions corresponding to $D_k$.

$$f_k(\mathbf{x}) = \sum_{i=1}^{N} \alpha_i y_k^{(i)} K\left(\mathbf{x}^{(i)}, \mathbf{x}\right) + b \quad (12)$$

In classification phase, we use the following rule to identify the class for input x.

$$k = \arg\max_k \left( f_k(\mathbf{x}) \right) \quad (13)$$

## 4. Implementation and Experiments

We select FERET and AT&T databases to evaluate our approach. The FERET database [12] was collected at George Mason University between August 1993 and July 1996. It contains 1564 sets of images for 14,126 images that include 1199 individuals and 365 duplicate sets of images. In our experiments, face regions of FERET images were identified and extracted from the background of the input images using the ground truth information of images but some images do not contain information on face locations. In this case, we used the well-known algorithm developed by Viola and Jones [13, 14] to find face positions. Then, they were scaled to 50-by-50 resolution. In dataset building task, we constructed a dataset D containing 1000 individuals which are chosen from sets fa, fb, fc, dup1 and dup2 of 1996 FERET database. All images of the dataset D are frontal face images. Next, we randomly divided the dataset into 3 separate subsets A, B and C. The reported results were obtained with Cross-Validation analysis on these subsets. We also use training set M of database provided by FERET for PCA feature extraction and 2DPCA extraction.

The AT&T database was taken at AT&T Laboratories. It contains 400 images (92-by-112) of 40 individuals; each person has ten images. We performed the same tasks to build datasets for experiments.

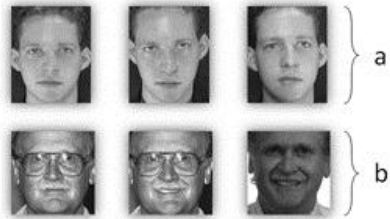

**Figure 5. a) Three faces from AT&T b) Three processed faces from FERET**

**4.1. Experiments on AT&T database**

We implemented five methods to conduct experiments on the AT&T database:

- MLP (PCA): This method uses PCA to extract feature vectors and Multi Layer Perceptron (MLP) for classification. The MLP has three layers: input layer has 163 nodes, hidden layer has 100 nodes, and output layer has 40 nodes. This MLP uses Gradient Back-Propagation algorithm for training. The active function of MLP is sigmoid function $f(x)$ and the range of learning rate $\eta$ is between 0.3 and 0.7.

$$f(x) = \frac{1}{1+e^{-x}}, f \in [0,1] \qquad (14)$$

- k-NN (PCA): We use PCA to obtain feature vectors and employ k-Nearest Neighbor (k-NN) with distance metric L2 for classification.

- SVM (PCA): It uses PCA to get feature vectors and applies SVM with two kernel functions (Polynomial, Radial Basis Functions-RBF) for classification. The value of $d$ of Polynomial is 3; for RBF kernel we used some values $C = \{2^{-5},...,2^{14}\}$ and $\sigma = \{2^{-15},...,2^{3}\}$ for classification.

$$K(\mathbf{x},\mathbf{x}') = (\mathbf{x}^T\mathbf{x}'+1)^d$$
$$K(\mathbf{x},\mathbf{x}') = e^{-\left(\frac{\|\mathbf{x}-\mathbf{x}'\|^2}{2\sigma^2}\right)} \qquad (15)$$

- k-NN (2DPCA): The method uses our proposed 2DPCA to get feature vectors and employs k-NN for classification.

- SVM (2DPCA): It uses the proposed 2DPCA to get feature vectors and SVM for classification.

We used the subset M to create PCA feature extractor. The default dimension of feature vector is k=163. With this k, we can get to a reasonable PCA reconstruction error of MSE = 0.0015. We also used the same subset M to create 2DPCA feature extractor. A weight for each training image is its rotate angle. The dimension of feature vector is k = 20.

For each method, we conducted three experimental trials on subsets A, B and C. It means that we trained classifiers on two subsets and evaluated on the remaining subset. The results are reported on their average performance scores in Table I.

The cumulative match score vs. rank curve for each method has been show in Fig. 6. The values of curve are the percentage of correct matches in the top n matches (rank-5).

The experimental results point that our proposed method for feature extraction is better than PCA and 2DPCA methods. As mentioned above, PCA is a method to reduce the dimension. There is not any mathematical evidence that it will increase the recognition rate. Our method has more advantages than traditional 2DPCA because it can create a subspace that reserves some importance discriminative information of face images such as pose.

The experimental results also show that MLP is the worst classification method and SVM is the best one. Obviously, MLP is easy to be overfitting because they usually focus on finding the lowest error rate although we use some techniques such as cross validation to limit the weak point. In other hand, SVM method always gives a suitable solution.

TABLE I. **EXPERIMENT RESULTS ON AT&T DATABASE**

| **Feature extraction** | **Classification** | **Accuracy (%)** |
|---|---|---|
| PCA | MLP | 75.2 |
| PCA | k-NN | 95.2 |
| PCA | SVM | 95.7 |
| 2DPCA | k-NN | 96.2 |
| 2DPCA | SVM | 97.3 |

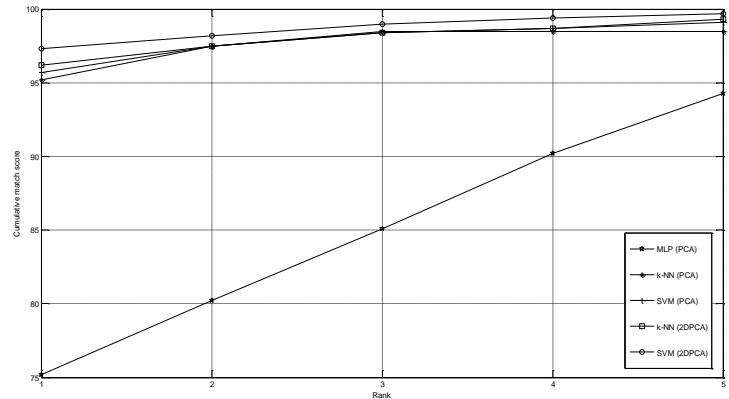

**Figure 6. Identification Performance on AT&T Database**

### 4.2. Experiments on FERET Database

We implemented four methods to conduct experiments on FERET database, which are k-NN (PCA), SVM (PCA), k-NN (2DPCA) and SVM (2DPCA). We did the same task to build feature extractors. First, we used the subset M to create PCA feature extractor. The default dimension of feature vector is k=100. Then, we continued to use the same subset M to create 2DPCA feature extractor. In our experiments, we set weight for female is 3, for male is 2 and for individual with glass is 1. It means that an image be easy to recognize has higher weight. The dimension of feature vector is k = 10.

We conducted three experimental trials on subsets A, B and C for each method. The results are reported on their average performance scores in Table II; and the cumulative match score vs. rank curve (rank-50) for each method has been shown in Fig. 7. The method 2DPCA with SVM for classification still gets the best performance on the FERET dataset.

TABLE II. **EXPERIMENT RESULTS ON FERET DATABASE**

| Feature extraction | Classification | Accuracy (%) |
|---|---|---|
| PCA | L2 | 80.1 |
| PCA | SVM | 85.2 |
| 2DPCA | L2 | 90.1 |
| 2DPCA | SVM | 95.1 |

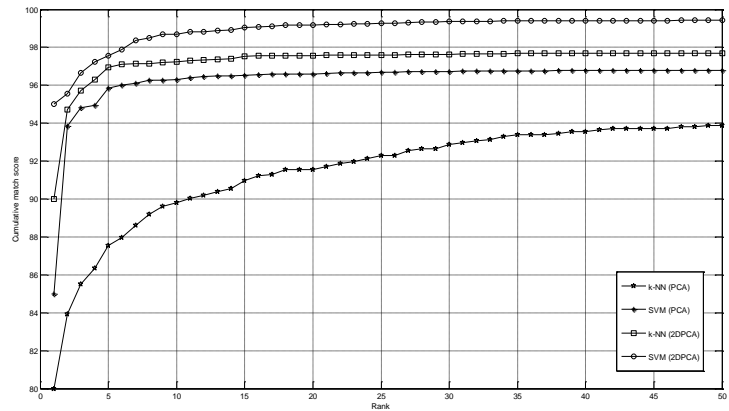

**Figure 7. Identification Performance on FERET Database**

## 5. Conclusions

In summary, we have proposed a new approach for face recognition. The first contribution of this paper is to propose a novel face model based on conventional 2DPCA for extracting feature vectors. The second contribution of this paper is to combine our proposed face model with SVM. We have compared our method with traditional methods. The results from our methods outperformed significantly.

## Authors

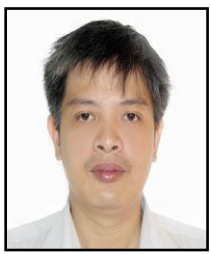

**Thai Hoang Le** received B.S degree and M.S degree in Computer Science from Hanoi University of Technology, Vietnam, in 1995 and 1997. He received Ph.D. degree in Computer Science from Ho Chi Minh University of Sciences, Vietnam, in 2004. Since 1999, he has been a lecturer at Faculty of Information Technology, Ho Chi Minh University of Science, Vietnam. His research interests include soft computing pattern recognition, image processing, biometric and computer vision. Dr. Thai Hoang Le is co-author over twenty five papers in international journals and international conferences.

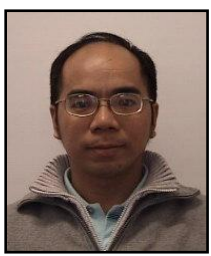

**Len Bui** received B.S degree and M.S degree in Computer Science from Ho Chi Minh University of Science, Vietnam, in 1996 and 2001. Since 2008, he has been a Ph.D. student in Computer Science at Faculty of Information Sciences and Engineering, University of Canberra, Australia.